\documentclass[sigconf]{acmart}

\usepackage{amsmath}
\usepackage{graphicx}
\usepackage{color}
\usepackage{booktabs}
\usepackage{multirow}
\usepackage{makecell}
\usepackage{multicol}
\usepackage{amsfonts}
\usepackage{textcomp}
\usepackage{algorithmic}
\usepackage[normalem]{ulem}
\usepackage[tight,footnotesize,center]{subfigure}
\usepackage{hyperref}
\AtBeginDocument{%
  \providecommand\BibTeX{{%
    \normalfont B\kern-0.5em{\scshape i\kern-0.25em b}\kern-0.8em\TeX}}}

\acmSubmissionID{1926}

\begin{document}

\title{Prompt Customization for Continual Learning}


\author{Yong~Dai}
\affiliation{%
  \institution{Pengcheng laboratory}
  \country{China}}
\email{chd-dy@foxmail.com}

\author{Xiaopeng~Hong}
\affiliation{%
  \institution{Harbin Institute of Technology}
  \country{China}
  }

\email{hongxiaopeng@hit.edu.cn}

\author{Yabin~Wang}
\affiliation{%
  \institution{Xi’an Jiaotong University}
  \country{China}
}

\author{Zhiheng~Ma}
\affiliation{%
  \institution{Shenzhen Institute of Advanced Technology, Chinese Academy of Sciences}
  \country{China}
}

\author{Dongmei~Jiang}
\affiliation{%
  \institution{Pengcheng laboratory}
  \country{China}
}
\author{Yaowei~Wang}
\affiliation{%
  \institution{Pengcheng laboratory}
  \country{China}
}



\begin{abstract}
  Contemporary continual learning approaches typically select prompts from a pool, which function as supplementary inputs to a pre-trained model. However, this strategy is hindered by the inherent noise of its selection approach when handling increasing tasks. In response to these challenges, we reformulate the prompting approach for continual learning and propose the prompt customization (PC) method. PC mainly comprises a prompt generation module (PGM) and a prompt modulation module (PMM). In contrast to conventional methods that employ hard prompt selection, PGM assigns different coefficients to prompts from a fixed-sized pool of prompts and generates tailored prompts. Moreover, PMM further modulates the prompts by adaptively assigning weights according to the correlations between input data and corresponding prompts. We evaluate our method on four benchmark datasets for three diverse settings, including the class, domain, and task-agnostic incremental learning tasks. Experimental results demonstrate consistent improvement (by up to 16.2\%), yielded by the proposed method, over the state-of-the-art (SOTA) techniques. The codes are released on \url{https://github.com/Yong-DAI/PC}.
\end{abstract}





\maketitle

\section{Introduction}
\label{sec:intro}
Continual learning pertains to the seamless integration of new learning tasks into a unified model while preventing catastrophic forgetting of previously acquired information~\cite{tcs3cong2023self,zhang2023slca,pivotvilla2023,tcs1lin2022anchor}. Continual learning has been involved in numerous applications including image classification, audio-visual learning, and vision-language pretraining in the image, multimedia, and multimodal domains \cite{jung2023generating,pointcloud,2021An,cheng2018deep,heng2023selective,mo2023classincremental,chen2024llmassisted,zhu2023ctp}. Early continual learning methods \cite{Regular-1,Regular-2} safeguarded entrenched knowledge against new information
through the application of regularization-based constraints to new task loss functions. The efficacy of these strategies was substantially influenced by the interplay between former and latter tasks. Subsequent methods \cite{rehearsal-3,rehearsal-4,LRCIL} have endeavored to retain a part of representative old data to reinforce previous knowledge during new task learning. Nonetheless, using data buffers has issues in data privacy and memory constraints \cite{prabhu2023computationally}. \par

\begin{figure}[!t]
	\centerline{\includegraphics[width=1\columnwidth]{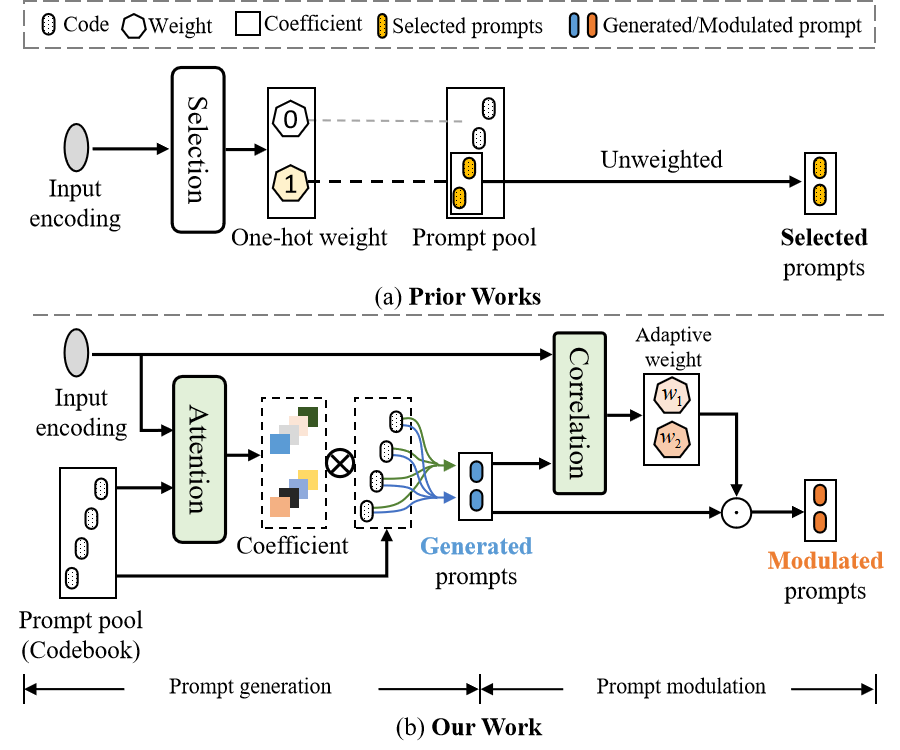}}
	\caption{ \emph{Prior works} \cite{L2P,dualprompt,Sprompts} employ a deterministic selection of prompts. \emph{Our work} tailors specific prompts for each instance through a process of \emph{prompt generation} from a fixed-size codebook and \emph{soft prompt modulation}. The key differences lie in the generated prompts by a linear combination of prompts based on instance-specific coefficients and modulated prompts by further assigning adaptive weights.}
	\label{fig:compare}
\end{figure}

Currently, top-performing methods are prompt-based methods \cite{L2P,Sprompts,dualprompt, coda}. Prompt learning is regarded as a flexible way to adapt models by solely training additional inputs while
keeping the model frozen. As Fig.~\ref{fig:compare}(a) shows, they usually learn to directly select prompts from a prompt pool. However, the performance of the above methods is highly dependent on this hard selection strategy, which becomes increasingly erratic as the number of tasks and the size of the prompt pool grows~\cite{gao2023unified}.
Moreover, these methods typically push to select the same prompt package for instances from a common task. Treating each task as a cohesive unit poses challenges due to the significant variability observed within tasks. In addition, it is also hard to provide more suitable instructions for each instance, which leads to prompts homogeneity problems \cite{gao2023unified}. \par

\begin{table}[!t]
	\centering
	\caption{Consistent performance Increase (Inc.) in terms of average accuracy ($A_a$, \%) using a simple prompt weighting operation on L2P \cite{L2P}. }
    \begin{tabular}{p{1.1cm}<{\centering}|p{1.4cm}<{\centering}p{1.6cm}<{\centering}p{0.9cm}<{\centering}p{1.5cm}<{\centering}}
    \toprule
    Datasets & \makecell[c]{Split\\CIFAR-100} & \makecell[c]{Split\\ImageNet-R} & CORe50 & DomainNet\\
    \midrule
    Inc. & 1.90 ${\uparrow}$ & 3.90 ${\uparrow}$ & 1.51 ${\uparrow}$ & 6.20 ${\uparrow}$\\
    \bottomrule
    \end{tabular}%
	\label{tab:plusw}%
\end{table}%

To handle the above problems caused by \emph{hard prompt selection}, we reformulate the prompting approach for continual learning. In contrast to conventional methods that employ hard prompt selection, as Fig.~\ref{fig:compare}(b) shows, we propose to generate instance-specific prompts, in which the generation strategy eliminates the selection stage and instance-specific prompts ease the homogeneity problems. The instance-specific prompt generation is achieved by assigning specific soft weights (coefficients) to prompts from a fixed-sized pool of prompts (codebook), which are then linearly combined.  Additionally, in our investigation, we have noted that even simple prompt weighting\footnote{The detailed structure is given in supplementary materials.} leads to consistent performance improvement, as evidenced in Tab.~\ref{tab:plusw}. This observation further catalyzes the design of an adaptive prompt modulation scheme to accommodate the generated prompts with the continually evolving codebook.\par

Based on this understanding, we propose the prompt customization (PC) method, which comprises two key modules: a prompt generation module (PGM) and a prompt modulation module (PMM). PGM adapts a predefined codebook to individual instances and generates corresponding prompts based on the predicted generation coefficient vectors derived from attention mechanisms between inputs and the codebook. Following this step, PMM adaptively modulates these generated prompts by assigning weights predicted by their correlations between inputs and respective prompts. The entire model is optimized using a straightforward yet highly effective loss function. Experimental results on four mainstream datasets
demonstrate that the proposed method outperforms the SOTA techniques significantly (up to 16.2\%) across diverse tasks, including class, domain, and task-agnostic incremental tasks.\par

The contributions of the proposed approach are threefold: 
\begin{itemize}
\item We propose prompt customization, a novel prompting method for continual learning. The proposed method leads to more variation and less homogeneity for prompts than hard selection and circumvents the need for prompt selection during inference, thereby mitigating the potential errors for prompt selection when dealing with challenging samples. 
\item We design a prompt generation module that generates finer instance-specific prompts through a linear combination of prompts from a designated codebook. The generated prompts are more distinguishable and expressive than generic task-specific prompts used in previous methods. 
\item We devise a prompt modulation module that further modulates the corresponding prompts with adaptive weights capitalizing on the correlations between instances and prompts and makes prompts more flexible.

\end{itemize}





\section{Related Work}
\label{sec:related-work}
\subsection{Continual Learning}
The field of continual learning has developed quickly in recent years, leading to the development of numerous methods \cite{tcs1lin2022anchor,tcs3cong2023self}. The landscape of architecture-based methodologies is characterized by their propensity to extract features from numerous intermediate layers \cite{54archi, Structbased-1, Structbased-2, Structbased-3}, or to extend models and fine-tune classification layers to learn new tasks \cite{Structexpand-1, Structexpand-2, Structexpand-3, rpsnet, dynaer, incrementershang2023}. However, their efficacy is curtailed due to the significant forgetting of preceding tasks, coupled with the expansion in network complexity \cite{rpsnet, dynaer}.\par

In contrast, regularization-based methodologies operate by inculcating constraints into the loss functions of new tasks, thereby averting the subjugation of previous knowledge by novel information \cite{Regular-1, Regular-2, LWF, EWC}. Although the methods seem to be potential, these constraints are markedly reliant on the interrelation between old and new tasks, rendering them less suitable for scenarios involving plenty of incremental tasks or intricate data distributions in challenging datasets \cite{dualhsicwang2023}.\par

The contemporary favor towards rehearsal-based techniques is underscored by their setting of preserving a subset of representative data of previous tasks during the training of new tasks, facilitating the revisitation of prior knowledge \cite{ER, Bic, Der++, Co2l, rehearsal-3, rehearsal-4, Gdumb, promptfusion}. Nevertheless, these strategies evince a pronounced susceptibility to the selection of previous data, a vulnerability that escalates linearly with the burgeoning task count \cite{ER, Bic, Der++, Co2l,tcs2_replay,tcs4select}.  However, these augmentations remain inadequate in circumventing concerns pertaining to data privacy and memory constraints \cite{privacy}.

\subsection{Prompt Tuning for Continual Learning}

Prompt tuning has garnered much attention as a versatile technique for model adaptation, involving the exclusive training of supplementary inputs in a parameter-efficient way while keeping the model frozen \cite{power35, visualprompt26, prefix38, coop, promptgn, zhoulearning76, liu2023pre}. This ingenious approach facilitates the model's effective reuse of learned representations, thereby circumventing the need for arduous and costly relearning from scratch \cite{visual3, Dytox, L2P, Sprompts, dualprompt, unsuperprompt24, coda, jung2023generating}.\par

\begin{figure*}[!t]
\centerline{\includegraphics[width=2\columnwidth]{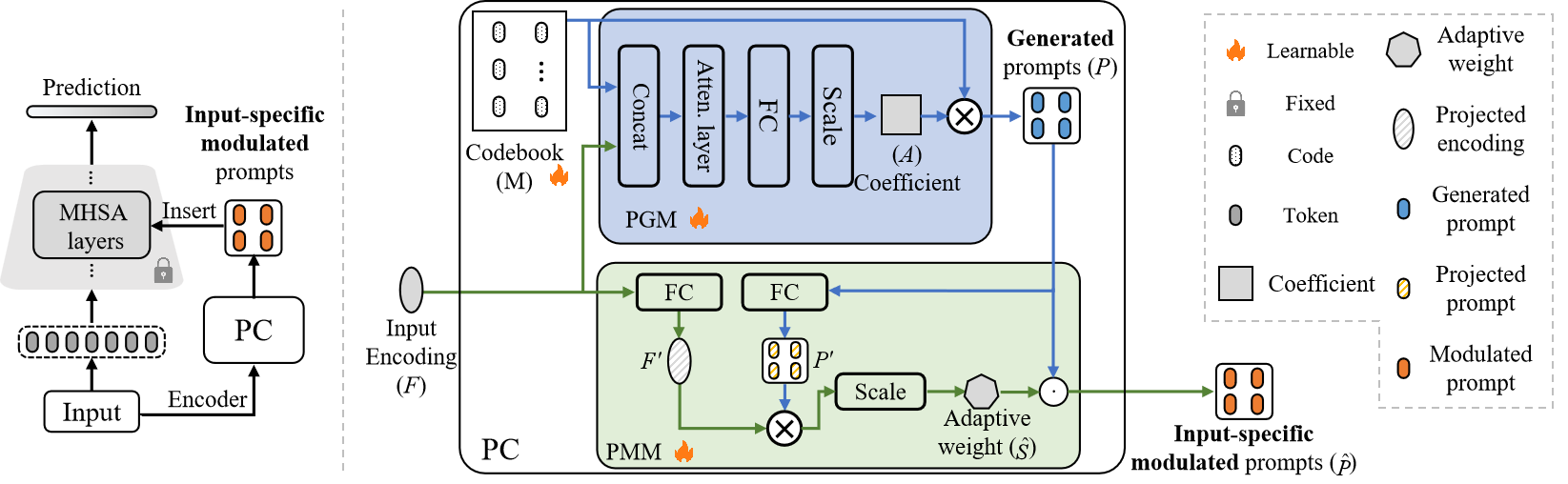}}
	\caption{The framework of the proposed PC. The PC comprises two integral modules: the PGM and PMM. PGM takes input encoding and codebook as input and generates input-specific prompts through a linear combination of prompts from a designated codebook. PMM takes input encoding and the above generated prompts as input to quantize their correlations which are further utilized to modulate the corresponding prompts. The final input-specific modulated prompts will be inserted into the corresponding MHSA layers to assist the frozen backbone in performing classification tasks. Unlike the previous works, our PC circumvents the need for rigorous prompt selection during inference and generates finer instance-specific prompts which are more distinguishable and expressive. `Atten.' and `MHSA' mean attention and multi-head self-attention layer, respectively. For simplicity, all symbols are present without subscripts.}
	\label{fig:method}
\end{figure*}
L2P stands as a pioneering endeavor in integrating prompts into the domain of continual learning, a stride marked by the elimination of rehearsal requirements. It introduces prompts as trainable parameters, enabling the direct selection of multiple prompts from a designated prompt pool \cite{L2P}. 
DualPrompt introduces a novel partitioning of the prompt pool into two distinct prompt spaces: task-agnostic and task-specific \cite{dualprompt}. A similar approach is adopted by S-Prompt, which independently learns task-specific prompts \cite{Sprompts}. This segregation of tuning strategies significantly mitigates the issue of catastrophic forgetting, albeit at the cost of prompt proliferation as the number of tasks increases. The prompts for the current task are directly selected based on an index, employing a one-hot weighting technique. However, this selection strategy is constrained by the growing noise due to the expanding prompt pool with the incorporation of additional tasks \cite{gao2023unified}. And the design of task-level prompts also limits the adaptivity of a model to each instance without finer instructions. \par
 
Moreover, CODA-Prompt \cite{coda} enhances the one-hot weighting mechanism by introducing an attention-based prompt-assemblage scheme, yielding promising performance. Nevertheless, it treats all the assembled prompts uniformly without further adjustment, which limits the adaptiveness of prompts to instruct the model. This motivates us to design the prompt modulation module to adaptively adjust the weight for each prompt. DAP \cite{jung2023generating} lies in the scope of generating instance-level prompts and achieves superior performance. However, it also involves a selection step to search for the domain-relevant knowledge of a target task, in which the superior performance is only obtained in the batch-wise selection setup, and the performance drops sharply in the instance-wise selection setup.


To address the previously delineated concerns, the proposed approach eschews the conventional prompt selection strategy. Instead, it undertakes the task of soft prompt generation and modulation on an instance-specific basis, facilitated by a predefined fixed-size codebook. This innovative approach circumvents issues associated with noisy selection and the linear proliferation of prompts and leads to more variation and less homogeneity for prompts.

\section{Proposed Approach}
\label{sec:approach}
\subsection{Problem Definition}
Assuming $D=\left\{ {{D_1},{D_2} \cdots {D_T}} \right\}$ be a sequence of tasks with ${T}$ incremental tasks, where ${t}$-th task ${D_{\rm{t}}} = \left\{ {\left( {x_i^t,y_i^t} \right)} \right\}_{i = 1}^{{n_t}}$ contains tuples of data sample $x_i^t \in X$ and the corresponding ground-truth labels $y_i^t \in Y$. The continual learning is defined to train a single model ${f_\theta }:X \to Y$ parameterized by $\theta$ to handle the ${T}$ incremental tasks. In the inferring phase, ${f_\theta }$ would predict the corresponding label $y \in Y$ for the given sample $x$ which is unseen from arbitrary tasks. Note that, the data from previous tasks may become inaccessible during the training of the current task.  \par

\subsection{Overall Framework}

In this work, we introduce a novel approach called PC, which caters to the customization of instance-level prompts. This process is facilitated by the novel soft generation and modulation of prompts, underpinned by a pre-established fixed-size codebook. Specifically, PC comprises two integral modules: the PGM and PMM. \par

As illustrated in Fig.~\ref{fig:method}, the PGM primarily generates tailored instance-specific prompts through a linear combination of prompts from a designated codebook. This generation is executed based on a coefficient vector, which is predicted through an attention mechanism integrating inputs and the codebook. Conversely, the PMM achieves adaptive prompt modulation by assigning dynamic weights to the generated prompts, capitalizing on the correlations between instances and prompts.\par

Subsequently, the modulated instance-specific prompts are seamlessly integrated into multiple layers of multi-head self-attention (MHSA). These integrated prompts culminate in the generation of final predictions, accomplished through a fixed Vision Transformer (VIT) as the backbone coupled with a learnable classifier.\par



\subsection{Prompt Generation Module}


We first create a learnable \textbf{codebook} in advance of prompt generation. This codebook is denoted as M which manifests as a matrix ${\mathbb{R}^{N \times L}}$ composed of $N$ fundamental codes of length $L$. 

With the codebook in place, the PGM is formulated to facilitate codebook adaptation for each task, achieved through the generation of instance-specific prompts guided by the coefficients. The concatenation of input encoding ${F_{i,t}}$ and the codebook M constitute the input for PGM, where ${F_{i,t}}$ stems from a previously frozen pre-trained vision model, predicated on the input data $x_i^t$. 
Leveraging attention layers, PGM establishes the link between ${F_{i,t}}$ and M, subsequently employing one FC layer and a scaling operation to predict and normalize the corresponding coefficient vector ${A_{i,t}}$ for the input data $x_i^t$. Finally, the corresponding generated instance-specific prompts ${P_{i,t}}$ are obtained through the matrix product of the coefficient vector ${A_{i,t}}$ and the codebook M as:
\begin{equation}
	\label{eq:PGM2P}
	{{P_{i,t}} = {A_{i,t}} \times {\rm M}}
\end{equation}
where the coefficient vector ${A_{i,t}}$ is with the size of ${\mathbb{R}^{n \times N}}$, so the corresponding generated prompts ${P_{i,t}}$ are with the size of ${\mathbb{R}^{{n} \times L}}$ which contains $n$ prompts of length $L$. \par

\subsection{Prompt Modulation Module}

Upon obtaining the $n$ instance-specific prompts via generation, a uniform allocation of contributions is present if the prompts are employed directly. This would imply an equal distribution of influence across the current instance prediction task. However, our perspective differs from this uniform allotment, believing that the expected distribution should be dynamically adjusted. Hence, we propose a modulation strategy based on the correlations between input encoding and the aforementioned $n$ prompts, which also makes the prompts adaptive.\par

Referencing Fig.~\ref{fig:method}, the PMM modulates the generated prompts through quantized correlations between input encoding and the generated prompts. In a precise breakdown, PMM initiates its operations by deploying a pair of fully connected layers with the projection dimension of ${\mathbb{R}^{L \times L}}$. This strategic configuration serves to delicately project the instance encoding ${F_{i,t}}$ and the instance-specific prompts ${P_{i,t}}$ into a shared representation space. The projections are respectively denoted as ${{F}'_{i,t}}$ and ${{P}'_{i,t}}$. Then PMM calculates their correlations by a matrix product operation as:

\begin{equation}
	\label{eq:PGM2Pmm}
	{{S_{i,t}} = {{F}'_{i,t}} \times {{P}'_{i,t}}}
\end{equation}

In order to quantize the correlations, PMM employs a scale operation by a sigmoid and a linear scale operation as:
\begin{equation}
	\label{eq:PGM2normS}
	{{{\hat S}_{i,t}} = \sigma (({{{S_{i,t}} - S_{i,t}^{\min })} \mathord{\left/
 {\vphantom {{{S_{i,t}} - S_{i,t}^{\min })} {(S_{i,t}^{\max } - S_{i,t}^{\min })}}} \right.
 \kern-\nulldelimiterspace} {(S_{i,t}^{\max } - S_{i,t}^{\min })}})}
\end{equation}
\noindent where ${{\hat S}_{i,t}}$ is the quantized correlations, $\sigma $ means the sigmoid non-linear operation. ${S_{i,t}^{\max }}$ and ${S_{i,t}^{\max }}$ mean the maximum and minimum value of ${{S_{i,t}}}$. The linear scale quantizes the correlations between 0 and 1. We hold the view that the quantized correlations for the generated prompts should not be too small, hence the sigmoid operation is further set after the linear scale operation to achieve non-linear operation and make the quantized correlations bigger than 0.5. Finally, PMM modulates the prompts according to the quantized correlations:

\begin{equation}
	{{{\hat P}_{i,t}} = {{\hat S}_{i,t}} \cdot  {P_{i,t}}}
\end{equation}

\subsection{Update Strategy}
The codebook serves as a universal and overarching foundation for all instance-specific prompts, effectively consolidating pivotal insights gleaned from incremental tasks. In order to aggregate key information to alleviate catastrophic forgetting, we suppose that the codebook ought to undergo stable updates. Specifically, we employ a momentum optimization~\cite{meanteacher}. We define ${\rm M_{t-1}}^\prime$ at task $t-1$ as an ensemble of the current version at task $t-1$ and earlier versions of codebook M:

\begin{equation}
	{{{\rm M}'_{t-1}}  = \alpha {{\rm M}'_{t - 2}}  + \left( {1 - \alpha } \right){{\rm M}_{{t-1}}}}
\end{equation}
where ${{\rm M}_{{t-1}}}$ is the learned codebook at task $t-1$, $\alpha$ is a smoothing coefficient which is set at 0.99 referring to \cite{meanteacher}. We further use a regularization item to further constrain the update of ${{\rm M}_{{t}}}$ under the guidance of ${{\rm M}'_{t-1}}$, thus, to encourage the ${{\rm M}_{{t}}}$ to aggregate key information of past tasks to alleviate catastrophic forgetting as:
\begin{equation}
	{{\mathcal L_{re}} =\frac{1}{N}\left\| {{{\rm M}'_{t - 1}}  - {{\rm M}_{\rm{t}}}} \right\|_2^2 }
\end{equation} 
What's more, to reduce interference between
each code in the codebook, we also add an orthogonality constraint as:
\begin{equation}
{{\mathcal L_{or}} ={\left\| {\rm{M}\rm{M^T} - I} \right\|_2} }
\end{equation} 

In contrast to conventional methods that employ a hard prompt selection approach and utilize a matching loss to optimize the distance between selected prompts and input from a common task, such a matching loss is not suitable for the proposed PC as PC eliminates the need for prompt selection. As for PC, the classification task is optimized with the cross-entropy loss ${\mathcal L_{ce}}$ as:

\begin{equation}
	{{\mathcal L_{ce}} =  - \frac{1}{{{n_t}}}\sum\limits_{i = 0}^{{n_t} - 1} {y_i^t\log \left( {\hat y_i^t} \right)} }
\end{equation}
where $\hat y_i^t$ means final predicted labels for each input $x_i^t$. Thus, the overall loss function is computed as: 
\begin{equation}
	{{\mathcal L}= {\mathcal L_{ce}} + \lambda {\mathcal L_{or}}+ \beta {\mathcal L_{re}}
 }
\end{equation}
$\beta$ is a scalar to weight the regularization item, and $\beta$ is set at 1 here referring to \cite{meanteacher}. The loss function is straightforward yet highly effective which is validated through extensive experiments.\par

\section{Experimental Setting and Results}
\label{sec:Experiments}
\begin{table*}[!t]
	\centering
	\caption{Performance comparison on the split CIFAR-100 dataset for class incremental learning setting. ${B_s}$ means buffer size. $^{\star}$ suggests results copied from the original paper, $^{\dagger}$ suggests results copied from (\cite{dualprompt}), $^{\ddagger}$ suggests results using the corresponding codebases and calculating by the normal equation. Results for the prompt-based methods are in \emph{instance-wise} prompts setup.}
	\begin{tabular}{p{2.4cm}|p{1.7cm}<{\centering}|p{1.6cm}<{\centering}p{1.6cm}<{\centering}|p{1.6cm}<{\centering}p{1.6cm}<{\centering}|p{1.6cm}<{\centering}p{1.6cm}<{\centering}}
    \toprule
    \multirow{2}[2]{*}{Method} & \multirow{2}[2]{*}{${B_s}$} & \multicolumn{2}{c|}{5 Tasks} & \multicolumn{2}{c|}{10 Tasks} & \multicolumn{2}{c}{20 Tasks}  \\
          &       & ${A_a}$ ${\uparrow}$  & ${F}$ ${\downarrow}$ & ${A_a}$ ${\uparrow}$  & ${F}$ ${\downarrow}$ & ${A_a}$ ${\uparrow}$  & ${F}$ ${\downarrow}$\\
    \midrule
    {DER++$^{\dagger}$ \cite{Der++}} & 1000  &   -    &   -    & 61.06 & 39.87 &   -    &   -    \\
   {BiC$^{\dagger}$ \cite{Bic}}   & 1000  &     -  &    -   & 66.11 & 35.24 &   -    &    -   \\
   {ER$^{\dagger}$ \cite{ER}}    & 1000  &    -   &    -   & 67.87 & 33.33 &   -    &  -     \\
   {C$o^2$L$^{\dagger}$ \cite{Co2l}}  & 1000  &     -  &     -  & 72.15 & 28.5  &   -    &   -    \\
   \midrule
    {EWC$^{\dagger}$ \cite{EWC}}   & 0     & -     & -    & 47.01 & 33.27 & -     & -    \\
   {LwF$^{\ddagger}$ \cite{LWF}}   & 0     & 61.55 & 25.13 & 60.69 & 27.77 & 56.46 & 28.87  \\
   \midrule
    {L2P$^{\ddagger}$ \cite{L2P}}   & 0     & 83.44 & 6.11  & 82.72    & 7.19  & 79.75 &  7.61 \\
    {DAP$^{\ddagger}$ \cite{jung2023generating}}   & 0     &    -   &    -   & 83.26 & 8.27  &   -    &    -   \\
    {DualPrompt$^{\ddagger}$ \cite{dualprompt}}& 0     & 85.92 & 5.62 & 85.33  & 5.71  & 82.38 & 6.11 \\
    {ESN$^{\star}$ \cite{wang2023isolation}}   & 0     & -     & -     & 86.34 & \textbf{4.76} & -     & -     \\
    {Coda-P$^{\ddagger}$ \cite{coda}} & 0     & 87.14 &  6.02 & 86.41 & 7.17  & 83.77 &  7.78  \\
    PC    & 0     & \textbf{88.04} &  \textbf{4.66} & \textbf{87.20} & 5.61  & \textbf{84.65} & \textbf{6.03}  \\
    \midrule
    Upper-bound & 0     &    90.97   &    -   & 90.97 & -     &    90.97   &    -  \\
    \bottomrule
    \end{tabular}%
	\label{tab:expe_CIL_C}%
\end{table*}%

\begin{table*}[!t]
	\centering
	\caption{Performance comparison on the ImageNet-R dataset for class incremental learning setting. ${B_s}$ means buffer size. $^{\star}$ suggests results copied from the original paper, $^{\dagger}$ suggests results copied from (\cite{dualprompt}), $^{\ddagger}$ suggests results using the corresponding codebases and calculating by the normal equation. Results for the prompt-based methods are in \emph{instance-wise} prompts setup.}
	\begin{tabular}{p{2.4cm}|p{1.6cm}<{\centering}|p{1.6cm}<{\centering}p{1.6cm}<{\centering}|p{1.6cm}<{\centering}p{1.6cm}<{\centering}|p{1.6cm}<{\centering}p{1.6cm}<{\centering}}
    \toprule
    \multirow{2}[2]{*}{Method} & \multirow{2}[2]{*}{${B_s}$} & \multicolumn{2}{c|}{5 Tasks} & \multicolumn{2}{c|}{10 Tasks} & \multicolumn{2}{c}{20 Tasks}  \\
          &       & ${A_a}$ ${\uparrow}$  & ${F}$ ${\downarrow}$ & ${A_a}$ ${\uparrow}$  & ${F}$ ${\downarrow}$ & ${A_a}$ ${\uparrow}$  & ${F}$ ${\downarrow}$ \\
    \midrule
    {DER++$^{\dagger}$ \cite{Der++}} & 1000  &    -   &   -    & 55.47 & 34.64 &   -    &   -    \\
    {BiC$^{\dagger}$ \cite{Bic}}   & 1000  &   -    &   -    & 52.14 & 36.7  &    -   &   -    \\
    {ER$^{\dagger}$ \cite{ER}}   & 1000  &    -   &   -    & 55.13 & 35.38 &   -    &   -    \\
    {C$o^2$L$^{\dagger}$ \cite{Co2l}}  & 1000  &   -    &   -    & 53.45 & 37.3  &   -    &   -    \\
    \midrule
    {EWC$^{\dagger}$ \cite{EWC}}   & 0     & -    & -     & 35.00    & 56.16 & -     & -     \\
    {LwF$^{\ddagger}$ \cite{LWF}}   & 0     & 40.62 & 50.69 & 38.54 & 52.37 & 32.05 & 53.42 \\
    \midrule
    {L2P$^{\ddagger}$ \cite{L2P}} & 0     & 62.61 & 8.01 & 61.21 & 8.65  & 57.36 & 9.07    \\
    {DualPrompt$^{\ddagger}$ \cite{dualprompt}} & 0     & 67.83 & \textbf{4.79 } & 66.47 & \textbf{5.75}  & 63.25 & \textbf{6.13}   \\
    {Coda-P$^{\ddagger}$ \cite{coda}} & 0     & 75.25 & 6.86 & 74.26 & 7.91  & 71.16 & 8.49   \\
    PC    & 0     & \textbf{75.41} & 6.42 & \textbf{74.34} & 7.35  & \textbf{71.44} & 7.62    \\
    \midrule
    Upper-bound & 0     & 79.31 &   -    & 79.31 &    -   & 79.31 &   -    \\
    \bottomrule
    \end{tabular}%
	\label{tab:expe_CIL_R}%
\end{table*}%

\begin{table}[!t]
	\centering
	\caption{Performance comparison on CORe50 for domain incremental learning setting. ${B_s}$ means buffer size. $^{\star}$ suggests results copied from original paper, $^{\dagger}$ and $^{\ddagger}$ suggest results copied from \cite{Sprompts} and \cite{promptfusion}, respectively.}
	\begin{tabular}{p{2.4cm}|p{2.0cm}<{\centering}|p{2.2cm}<{\centering}}
		\toprule
		{Method} & {${B_s}$} &{${A_a}$ ${\uparrow}$} \\
		\midrule
		{DER++$^{\dagger}$ \cite{Der++}} & {2500} & 79.70  \\
		{BiC$^{\dagger}$ \cite{Bic}} & {2500} & 79.28  \\
		{ER$^{\dagger}$ \cite{ER}} & {2500} & 80.10   \\
		{C$o^2$L$^{\dagger}$ \cite{Co2l}} & {2500} & 79.75  \\
		{L2P$^{\star}$ \cite{L2P}} & {2500} & 81.07  \\
		\midrule
		{EWC$^{\dagger}$ \cite{EWC}} & {0} & 74.82  \\
		{LwF$^{\dagger}$ \cite{LWF}} & {0} & 75.45  \\
		{L2P$^{\star}$ \cite{L2P}} & {0} & 78.33  \\
		{S-iPrompt$^{\star}$ \cite{Sprompts}} & {0} & 83.13  \\
		{DualPrompt$^{\ddagger}$ \cite{dualprompt}} & {0} & 87.20  \\
		{\textbf{PC}} & {0} & \textbf{91.35}  \\
		\midrule
		{Upper-bound} & {-} & 92.20  \\
		\bottomrule
	\end{tabular}%
	\label{tab:dilcore}%
\end{table}%

\begin{table}[!t]
	\centering
	\caption{Performance (task-agnostic domain-incremental Learning) comparison on DomainNet. ${B_s}$ means buffer size. $^{\star}$ suggests results copied from the original paper, $^{\dagger}$ suggests results copied from (\cite{Sprompts}), $^{\ddagger}$ suggests results using the corresponding codebases and calculating by the normal equation.}
	\begin{tabular}{p{2.4cm}|p{2.0cm}<{\centering}|p{2.2cm}<{\centering}}
		\toprule
		{Method} & {${B_s}$} & {${A_a}$ ${\uparrow}$} \\
		\midrule
		{EWC$^{\dagger}$ \cite{EWC}} & {0} & 47.62 \\
		{LwF$^{\dagger}$ \cite{LWF}} & {0} & 49.19 \\
		{L2P$^{\dagger}$ \cite{L2P}} & {0} & 40.15 \\
		{S-iPrompt$^{\star}$ \cite{Sprompts}} & {0} & 50.62 \\
		{DualPrompt$^{\ddagger}$ \cite{dualprompt}} & {0} & 49.30 \\
		{\textbf{PC}} & {0} & \textbf{58.82} \\
		\midrule
	{Upper-bound} & {-}   & 63.22 \\
		\bottomrule
	\end{tabular}%
	\label{tab:dildomainnet}%
\end{table}%

\subsection{Datasets and Experimental Preparation}

\subsubsection{Datasets}

In this work, we would like to verify the validity of the proposed PC in multiple aspects, including the class, domain, and task-agnostic incremental learning settings\footnote{The details are also given in supplementary materials.}. Therefore, we follow the prior work, i.e. L2P \cite{L2P}, SPrompt \cite{Sprompts}, and ESN \cite{wang2023isolation}, and pick the representative datasets for a fair comparison. For the class-incremental task, we consider the CIFAR-100 \cite{cifar100} and ImageNet-Rendition (ImageNet-R) \cite{ImageNet,dualprompt} datasets as the base datasets and randomly split them into 5, 10, and 20 incremental classification tasks respectively for each task. For the domain-incremental task, we conduct experiments on Core50 \cite{core50} to predict the unseen domains based on incremental trained domains. And for the task-agnostic incremental learning settings, we apply an incremental training process and present the task-agnostic test average results on DomainNet \cite{domainnet}. 

\subsubsection{Compared Methods}

Task identity is not available in the context of class-incremental learning, thus we resort to selecting the most established and high-performing methods without prior knowledge of task identity during the testing phase. The compared methods contain regularization-based methods (EWC \cite{EWC} and LwF \cite{LWF}), rehearsal-based methods (ER \cite{ER}, BiC \cite{Bic}, DER++ \cite{Der++} and C$o^2$L \cite{Co2l}), as well as the prompt-based methods (L2P \cite{L2P},S-Prompt \cite{Sprompts}, DualPrompt \cite{dualprompt}, CODA \cite{coda}, ESN \cite{wang2023isolation}, and DAP \cite{jung2023generating} ) . \par

 To make a fair comparison, the above methods as well as the proposed PC\footnote{Implementation details and training schedule of the proposed PC are given in the supplementary materials.} employ the pre-trained ViT-B/16 as the backbone. Specifically, the methods (L2P, DualPrompt, and DAP ) involving selection strategy usually present the inferring results in the batch-wise prompt setup where a single set of prompts is chosen for the entire batch. In this setup, the frequently used prompts are selected to correct potential errors when dealing with challenging samples, which is not realistic. We hold the view that results should be given in \emph{instance-wise} prompts setup where prompts are selected on a per-instance basis. Hence, in this paper, we give the performance for the prompt-based methods in \emph{instance-wise} prompts setup.

Note that, the upper bound is typically achieved through supervised fine-tuning on the i.i.d. data from all tasks, which is commonly considered as the best performance a continual learning method can attain. Besides, we utilize two widely-used metrics\footnote{The details are given in supplementary materials.}: Average Accuracy (${A_a}$, the higher, the better”) and Forgetting ($F$ the lower, the better) \cite{gradient2017}. Unless otherwise specified, we report the performance in \emph{percentile} and omit the subscript.

\subsection{Experimental Results}
\subsubsection{Results on Class-incremental Learning.}

Tab.~\ref{tab:expe_CIL_C} and Tab.~\ref{tab:expe_CIL_R} report the performance of the datasets of Split Cifar-100 and Split ImageNet-R. As for the common task setting (10 tasks), we observe that the proposed PC achieves 87.20\% and 74.34\% of ${A_a}$ measure, respectively, which achieves remarkable improvement over the regularization-based methods (an average of roughly 64.6\% and 102.6\% on Split Cifar-100 and Split ImageNet-R, respectively) and the rehearsal-based methods (an average of roughly 31.0\% and 37.6\% on Split Cifar-100 and Split ImageNet-R, respectively) even though they use many buffers. The proposed PC also outperforms all the compared prompt-based methods (an average of roughly 2.4\% and 7.0\% on Split Cifar-100 and Split ImageNet-R, respectively) in terms of ${A_a}$ measure.
The kinds of methods prove to be insufficiently effective due to the significant variability across tasks, thereby limiting the precision of prompt \emph{instance-wise} selection. Consequently, these limitations hinder the optimal performance they are supposed to achieve through their prompts-separate optimization.\par
Among these methods, only CODA-P\footnote{We implement the official codes of CODA-P and present the results of each task in supplementary materials.} performs comparably to our method. However, PC outperforms CODA-P by a margin of 0.8\% and 0.1\% in terms of average accuracy due to our instance-level design, which effectively mitigates intra-task variance and eliminates noisy selection strategies. Although the proposed PC achieves slightly higher Forgetting values on the two datasets and secures the second-best position, it remains highly competitive.\par

 In addition to the 10-task setting, we also provide results with other settings to compare the scalability of each method based on a smaller (5 tasks) and a larger number of tasks (20 tasks), respectively. The results show that the proposed PC achieves the best results in terms of ${A_a}$ measure no matter what the task setting is. And PC only decreases slowly when handling more tasks, which shows that our method scales stably to different task number settings.

\begin{table}[!t]
	\centering
	\caption{Performance comparison.  ${T_t}$ means the training time (hour).  $^{\ddagger}$ suggests results using the corresponding codebases.}

    \begin{tabular}{p{2.4cm}|p{2.0cm}<{\centering}|p{2.2cm}<{\centering}}
		\toprule
		{Method} & {${A_a}$} & {${T_t}$ ${\uparrow}$} \\
		\midrule
    L2P$^{\ddagger}$ \cite{L2P}  & 83.00    & 0.87 \\
    DualPrompt$^{\ddagger}$ \cite{dualprompt}&  85.80   & 0.94 \\
    CODA-P$^{\ddagger}$ \cite{coda} &  86.41  & 3.60 \\
    \textbf{PC-light} & 86.68  & 1.06 \\
    \textbf{PC}   & \textbf{87.20} & 1.78 \\
    \midrule
    Upper-bound & 90.85     & - \\
    \bottomrule
	\end{tabular}%
    
	\label{tab:expe_CIL_Arc}%
\end{table}%
\subsubsection{Results on Domain-incremental Learning}

We further consider the DIL tasks on CORe50 with the corresponding results shown in Tab.~\ref{tab:dilcore}. It is worth noting that none of the tested domains in CORe50 were included in the incremental training process. The Forgetting measure is not suitable for the unseen tested domain setting, thus only the results of $A_a$ are given. The rehearsal-based methods demonstrate superior performance compared to EWC, LwF, and even L2P, suggesting that incorporating large buffers may yield marginal improvements over supervised training. However, they still fall short of the exceptional performance achieved by prompt-based methods such as S-iPrompt, DualPrompt, and PC, highlighting the clear advantage of rehearsal-free approaches. S-iPrompt is a variant of S-Prompt, it achieves significant advances compared to L2P due to easing the inter-task interference problem of L2P. DualPrompt further improves the performance compared to L2P and S-iPrompt with the benefits of multi-layer prompts. Moreover, the limited inter-domain variance in the CORe50 dataset minimizes the disparities among prompts from different tasks, thereby promoting the superior performance of S-iPrompt and DualPrompt even when wrong prompts are selected. Nevertheless, our PC attains an impressive 91.35\% performance on the ${A_a}$ measure, a result that closely approaches the upper-bound performance threshold. Moreover, the proposed PC consistently outperforms all other compared methods, showcasing the exceptional generalization capabilities of our PC on previously unseen domains. This can be attributed to the utilization of a global and comprehensive codebook, as well as the corresponding prompt generation and modulation modules. 

\subsubsection{Results on Task-Agnostic Domain-incremental Learning}

For DIL tasks, the task-agnostic setting is generally regarded as more challenging \cite{shanahan2021encoders}. The Forgetting measure is also not suitable for the task-agnostic setting and thus is omitted. Experiments of the setting are conducted on DomainNet with the results shown in Tab.~\ref{tab:dildomainnet}. Note that, 345 classes are selected for each task in this experiment, which is highly challenging. The variance between domains in DomainNet is significantly larger compared to CORe50, which greatly benefits the precision of the selection strategy in L2P, S-iPrompt, and DualPrompt. 
However, the proposed PC still exhibits its superiority and achieves 58.82\% in terms of ${A_a}$ measure with an approximately 16.2\% improvement compared to the second best exemplar-free DIL method S-iPrompt. The results also demonstrate the effectiveness of the PMM in significantly enhancing the capacity of prompts to handle domain shifts.

\begin{table*}[!t]
	\centering
	\caption{Performance of ablation study. }
    \begin{tabular}{p{3.3cm}|p{1.8cm}<{\centering}p{1.8cm}<{\centering}|p{1.8cm}<{\centering}p{1.8cm}<{\centering}|p{1.8cm}<{\centering}|p{1.8cm}<{\centering}}
    \toprule
    \multirow{2}[2]{*}{Method} & \multicolumn{2}{c|}{Split Cifar-100} & \multicolumn{2}{c|}{Split ImageNet-R} & Core50 & DomainNet \\
          & {${A_a}$ ${\uparrow}$ }  & {${F}$ ${\downarrow}$} & {${A_a}$ ${\uparrow}$ }  & {${F}$ ${\downarrow}$} & {${A_a}$ ${\uparrow}$ }  & {${A_a}$ ${\uparrow}$ }  \\
    \midrule
    Baseline  & 69.93 & 8.44  & 53.55 & 9.20 & 71.28 & 36.54 \\
    +select prompts & 85.51 & 5.91  & 67.75 & 9.03  & 86.89 & 49.05 \\
    +PGM & 87.03 & 5.68  & 73.51 & 7.79  & 90.54 & 57.53 \\
    +PGM+spw & 87.12 & 5.65  & 73.67 & 7.65  & 90.83 & 57.56 \\
    \textbf{+PGM+PMM} & \textbf{87.20} & \textbf{5.61} & \textbf{74.34} & \textbf{7.35} & \textbf{91.35} & \textbf{58.82} \\
    \bottomrule
    \end{tabular}%
	\label{tab:ab_study}%
\end{table*}%

\subsection{Analysis of The Proposed Method}
\subsubsection{Training time}
To further demonstrate the effectiveness of the PC, we also report the results of prompt-based methods in terms of $A_a$ and training time ($T_t$) based on the split CIFAR-100 dataset\footnote{We train each method with the default setting referring to the official paper, detailed training setting and platform are given in the supplementary materials.}.  As Tab.~\ref{tab:expe_CIL_Arc} shows, the DualPrompt model surpasses L2P in terms of performance by employing prompts on multiple layers with only a little more training time. Additionally, CODA-P enhances the performance further by incorporating attention calculation based on DualPrompt. However, CODA-P requires almost 4 times the training time of DualPrompt according to the corresponding default settings. We propose to employ a dual-attention mechanism in the context of PC, aiming to generate and modulate prompts. Notably, even with a near training time, the light PC (PC-light) consistently outperforms DualPrompt by an absolute 0.88\% and even outperforms CODA-P by 0.26\% with only an appropriate quarter training time. Furthermore, increasing the number of attention layers in PC leads to additional advantages with an absolute 0.79\% improvement with only an appropriate half training time compared to CODA-P.

\subsubsection{Diversity of the prompts generated}
\begin{figure}[!t]
	\centering{ \subfigure[]{\includegraphics[width=1\columnwidth]{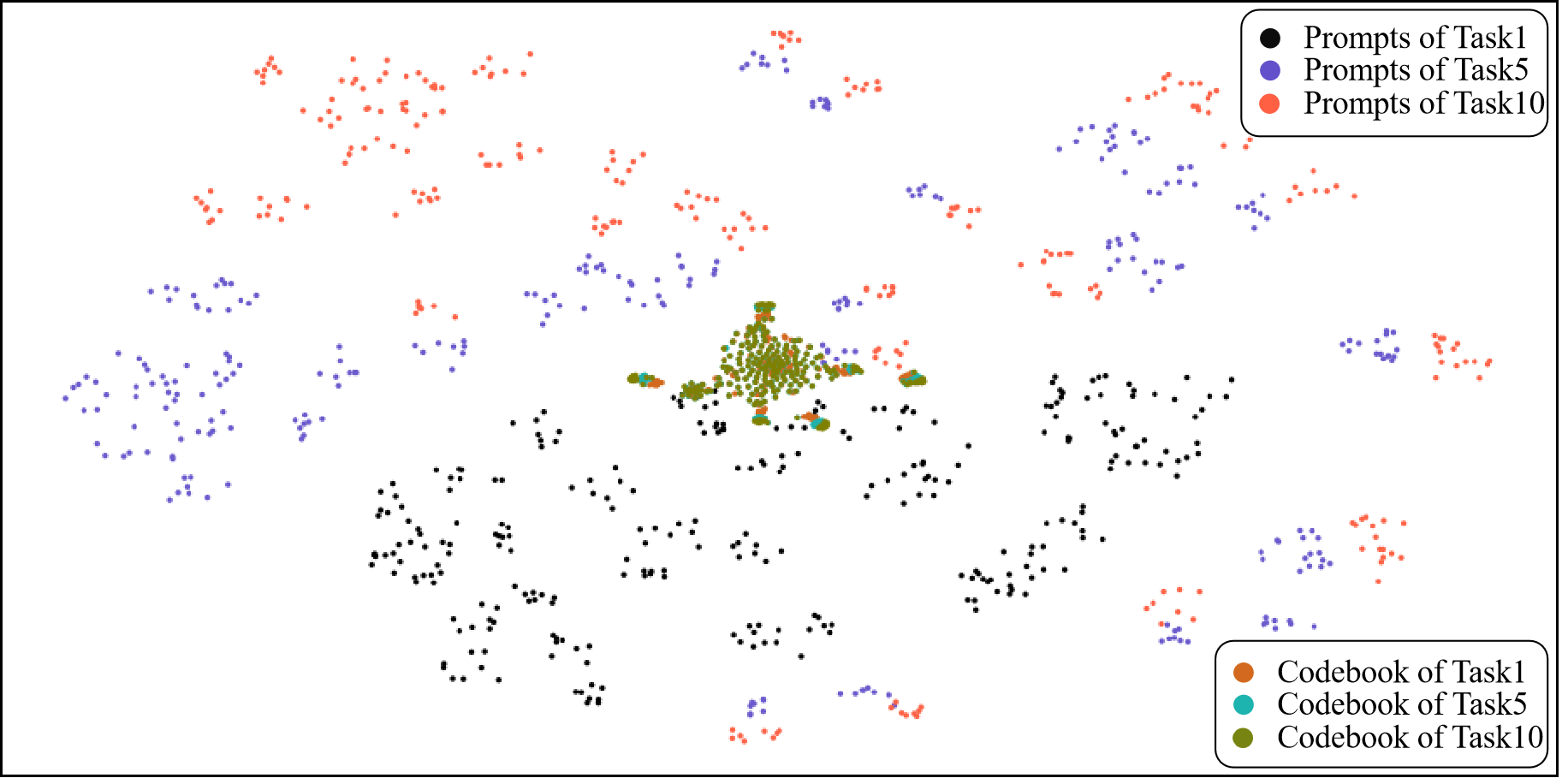}}
		\subfigure[]{\includegraphics[width=.9\columnwidth]{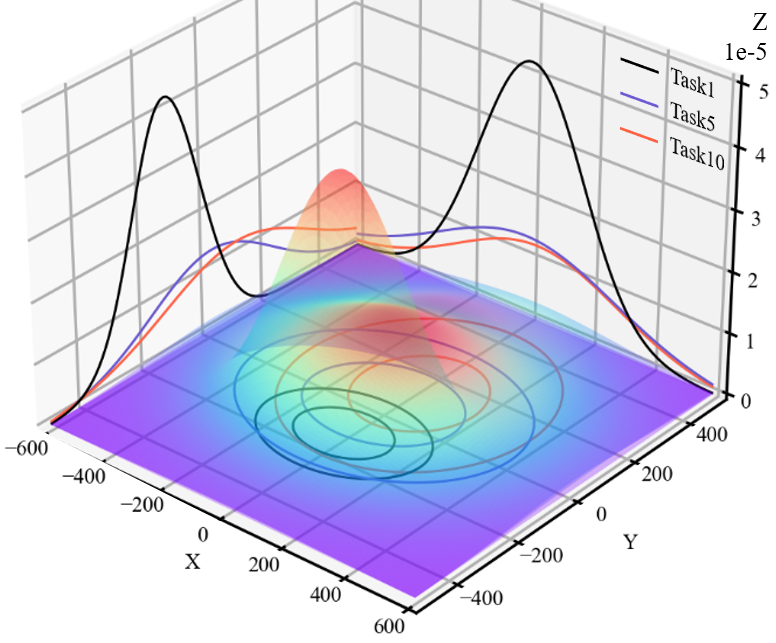}}}
	\caption{The visualization of codebook and prompts for the first, fifth, and tenth tasks. (a) t-SNE visualization of the updated codebook and prompts; (b) Gaussian distribution of prompts. The figure is best viewed in color.}
	\label{fig:homogeneity}
\end{figure}

As previously detailed in Section \ref{sec:approach}, the iterative augmentation of task numbers concurrent updates to both the codebook and prompts. We visualize the codebook and prompts corresponding to the initial, fifth, and tenth tasks by t-SNE~\cite{tsne} depicted in Fig. \ref{fig:homogeneity} (a). Additionally, Fig. \ref{fig:homogeneity} (b) portrays the 2D Gaussian distribution of the prompts' t-SNE projections for the corresponding tasks.

The center of Fig. \ref{fig:homogeneity} (a) depicts the codes in the codebook after the learning of tasks 1, 5, and 10. The remaining segments of Fig. \ref{fig:homogeneity} display the prompts of the test data after the completion of the training process. Our observations are threefold: Firstly, the prompts associated with each task exhibit a notably greater diversity compared to the codebook,  thus affirming the efficacy of the proposed prompt customization approach; Secondly, the iterative updates of the prompts across tasks are discernible, which enables the prompts are adaptive to the instances of different tasks. A more in-depth exploration by examining the 2D Gaussian distribution of the t-SNE projections of prompts is depicted in Fig. \ref{fig:homogeneity} (b). Remarkably, it is evident that the proposed PC adeptly tailors prompts across a varied range for each task, effectively managing the data distribution shifts; Thirdly, the update of the codebook remains relatively stable, primarily attributed to the momentum optimization with the imposed orthogonality constraint, forming a strong basis for forgetting alleviation.

\subsection{Effectiveness Analysis}
\subsubsection{Codebook Related learning}

\begin{figure}[!t]
	\centerline{\includegraphics[width=0.95\columnwidth]{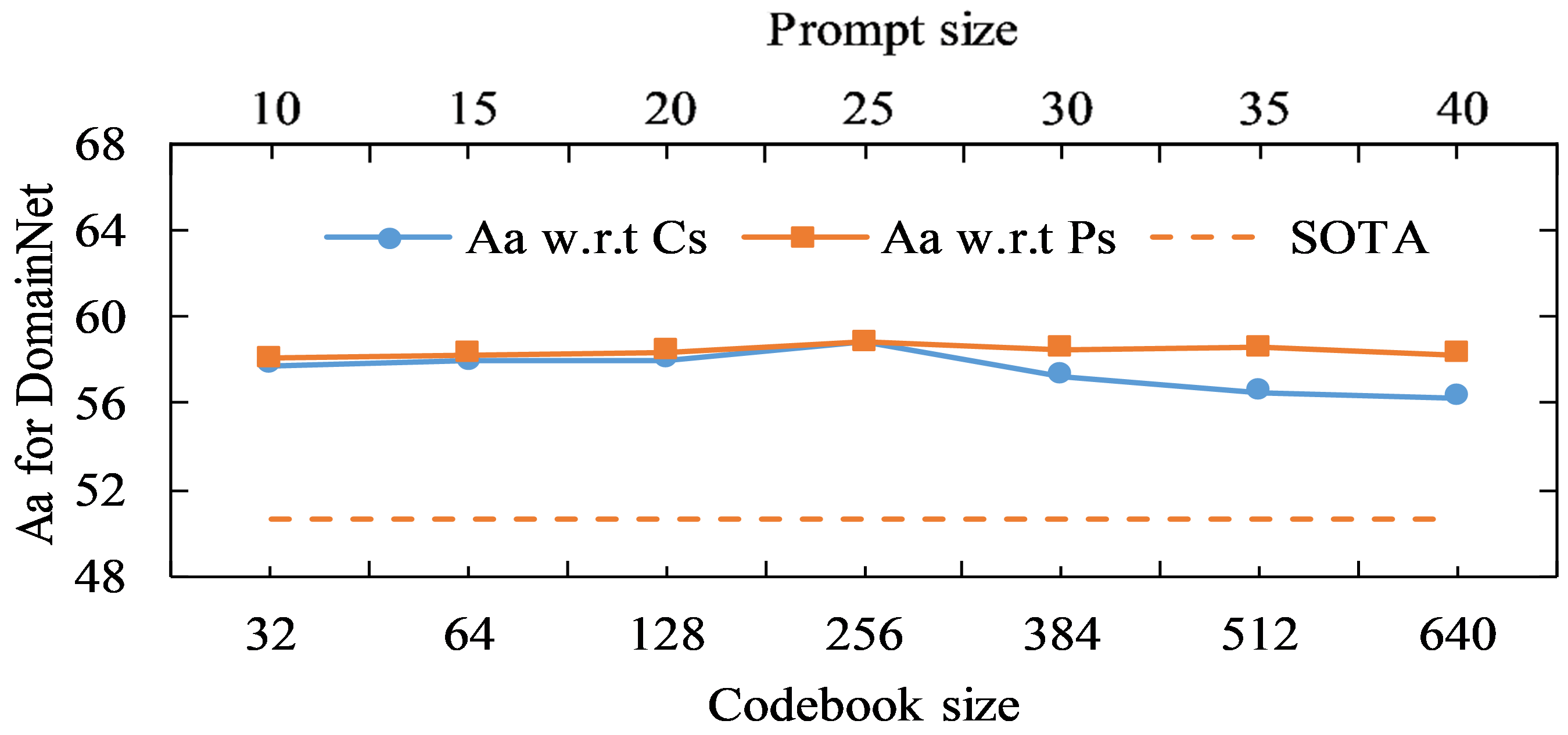}}
	\caption{The performance with regard to different codebook sizes (Cs) and prompt sizes (Ps). The results demonstrate the stability of the proposed PC with small fluctuations concerning different parameters.}
	\label{fig:codebook}
\end{figure}

The establishment of the codebook serves as a foundational element for prompt generation, playing a pivotal role in shaping the method's capacity to handle incremental tasks. Fig.~\ref{fig:codebook} offers insights into the impact of codebook size on outcomes. Our observations reveal that augmenting the codebook size from 32 to 256 yields a marginal improvement in performance. The most optimal results, however, materialize when the codebook size is set at 256.\par

Remarkably, our approach consistently delivers competitive performance even with a more modest codebook size of 32. This underscores the method's adaptability and its capacity to achieve commendable results even under such conditions.

\begin{figure}[!t]
	\centerline{\includegraphics[width=0.95\columnwidth]{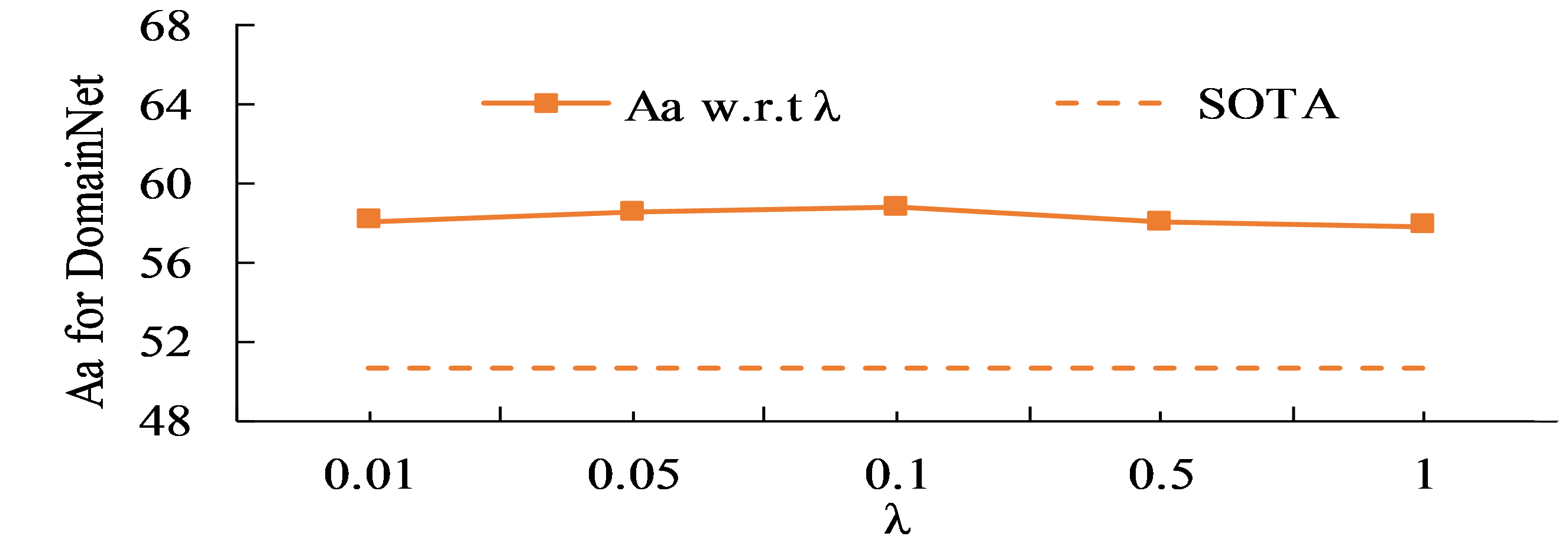}}
	\caption{The performance respects to the weight $\lambda$. }
	\label{fig:lambda}
\end{figure}

Moreover, to enhance the diversity of the codebook and mitigate interference among its components, an orthogonality constraint is introduced. The impact of its weight $\lambda$ on performance is depicted in Fig. \ref{fig:lambda}. The results illustrate that incrementally augmenting $\lambda$ up to 0.1 yields consistent enhancements in performance. However, as $\lambda$ is further increased, there is a concurrent reduction in the influence of the classification loss term, ${\mathcal L_{ce}}$, potentially leading to a detrimental effect on overall performance.

\subsubsection{Prompt Related Learning}
\label{sec:promplen}


The proposed PGM takes input and codebook as input to compute their attention and further linearly combines the codebook for the corresponding input. 

In the field of prompt-based methods, a common approach is to enlarge the size of the prompt to attain improved performance. However, as depicted in Fig.~\ref{fig:codebook}, our PC approach distinguishes itself by showcasing a notable degree of insensitivity to prompt size variations. This insensitivity highlights the robustness of the PC method, as it manages to maintain competitive performance levels even when the prompt size is reduced. This reduction in prompt size not only serves to enhance parameter efficiency but also ensures the method's capability to deliver commendable results.\par

In light of these findings, in this paper, we have determined an optimal prompt size of 25. This choice reflects a balance between maintaining effective performance and optimizing parameter efficiency within the context of the PC approach.\par

\subsubsection{Ablation Study}

The proposed method adopts prompt generation and modulation steps to obtain instance-specific prompts. The ablation studies in Tab.~\ref{tab:ab_study} justify the necessity of generation and modulation steps. Firstly, we set a baseline where the backbone is the same as that of the PC and we only update the final classification layer while keeping the backbone frozen. The performances (e.g., 69.93\% on Split Cifar-100) are with much forgetting (e.g., 8.44\% on Split Cifar-100) since the update process of prompts for the new tasks has a negative effect on the old tasks. Then we utilize the directly select strategy in L2P \cite{L2P} to select 25 prompts (``+selected prompts'') from the codebook, the corresponding results (e.g., 85.51\% on Split Cifar-100) improve a lot compared to the baseline which demonstrates the effectiveness of prompt tuning approach. However, the results are still not satisfied, this may be due to the inter-task interference problem by the noisy select strategy for each instance. This variant is similar to DualPrompt while this variant has a larger prompt pool. This variant is also not better than DualPrompt with a smaller prompt size, which is also imagined according to Fig.~\ref{fig:codebook} since increasing the prompt size has saturated or even negative returns. Next, the performance (e.g., 87.03\% on Split Cifar-100) is better than the above two variants when we only employ the PGM (``+PGM''), which demonstrates the effectiveness of the soft generation strategy of PGM compared to the direct selection strategy in prior works. What is more, we further add the simple prompt weighting operation (``+PGM+spw'') which still obtains a little performance improvement, as evidenced in Tab.~\ref{tab:plusw}. Finally, the difference between the variant ``+PGM'' and the proposed PC (``+PGM+PMM'') also shows that the PMM is able to increase the capacity of prompts, thus being effective in further improving performance. \par


\section{Conclusion}
\label{sec:conclusion}
In this paper, we propose a prompt customization method for continual learning. Specifically, we propose PGM and PMM techniques to enable prompt customization by generating and modulating instance-specific prompts. The proposed method replaces the hard selection of prompts with flexible and efficient prompt generation and modulation. The strategy leads to more variation and less homogeneity for prompts and avoids the inherent noise of the selection strategy when handling increasing tasks. We perform extensive experiments on four benchmark datasets across three distinct settings, encompassing class, domain, and more challenging task-agnostic incremental learning scenarios. The experimental results demonstrate the general applicability and robustness of the proposed method for incremental classes and even unseen domains. In the future, we will adapt the proposed method to other multimedia or multimodal datasets and explore more sophisticated and generalized methods for prompt generation without using a prompt pool or codebook, which holds promise for enhancing prompt adaptation across different modalities.






\appendix
\section{Appendices}
In this part, we additionally present the detailed framework of the {Plus\_W}, details setting of the benchmark datasets, thorough implementation details and train schedule of PC, other related analyses, and the detailed experiment results of CODA-P. The full implementation code has given in the supplementary materials and will be made publicly available.

\section{Datailed pipeline of {Plus\_W}}
\label{sec:Plus_W}

\begin{figure}[!t]
	\centerline{\includegraphics[width=1\columnwidth]{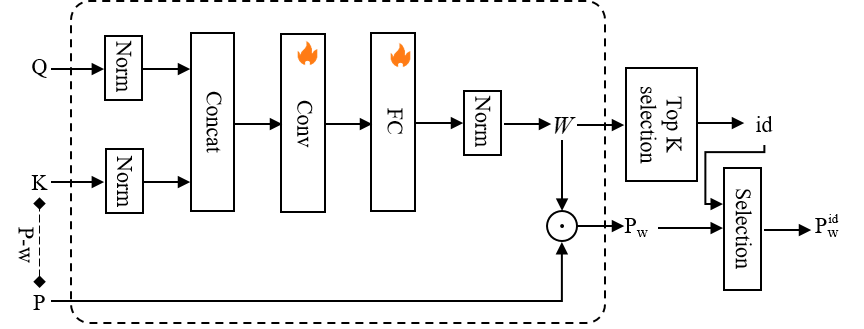}}
	\caption{The pipeline of {Plus\_W}. Q, K, and P mean query, keys, and prompts, respectively, given L2P. `P-w' denotes keys and prompts are pair-wise tuples. `Norm', `Concat' are normalization and concatenation. `Conv' and `FC' mean convolution and fully connected layers.}
\label{fig:pluswline}
\end{figure}

We investigate that a simple weighting operation for prompt boosts the corresponding performance to some extent. The weighting operation is given in Fig. \ref{fig:pluswline}. We first employ normalization to make the query and keys at the same scale. Then we concatenate them and employ a convolution operation to encourage the query to interact with each key. We further set a FC layer and a normalization operation to predict and normalize the corresponding weight $W$. The $W$ is then utilized to weight the prompts by a dot product. The weighted prompts are denoted as ${\rm P_w}$ which is then selected ($\rm {P_w^{{id}}}$) according to the top K indexes (Id) of ranked $W$. Note that, we calculate the weight between the query and keys instead of prompts, which decreases the potential negative effect to the prompt optimization. ${\rm Plus\_W}$ is a straightforward but effective prompt weighting method, leading to consistent performance improvement.

\section{Detailed setting of the benchmark datasets and evaluation matrices}

\subsection{Datasets}
In this work, for the class-incremental task, we consider the CIFAR-100 and ImageNet-Rendition (ImageNet-R) datasets as the base datasets and randomly split them into 10 incremental classification tasks with 10 classes or 20 classes respectively for each task. And for the domain-incremental task, we conduct experiments on Core50 and DomainNet. Details are given as follows.

\textbf{Split CIFAR-100}
The CIFAR-100 dataset has 100 classes, each containing 600 images, including 500 training pictures and 100 test pictures \cite{cifar100}. The 100 classes in CIFAR-100 are divided into 20 super-classes. Each image has a ``fine'' label (the class to which it belongs) and a ``rough'' label (the super-class to which it belongs). Dozens of related pieces of literature usually split the original CIFAR-100 according to the ``fine'' label into 10 separate tasks (Split CIFAR-100) which is regarded as a commonly used benchmark in the field of continual learning. It divides the original CIFAR-100  into 10 separate tasks, each containing 10 classes. Although it may seem like a relatively simple task for image classification under independent and identically distributed (i.i.d.) conditions, it is challenging and effectively exposes significant forgetting rates in class-incremental learning when advanced incremental learning methods are applied \cite{dualprompt}.

\textbf{Split ImageNet-R}
ImageNet-R is an extension of the ImageNet dataset \cite{ImageNet}, which includes various renditions of multiple styles such as art, cartoon, DIYART, graffiti, embroidery, graphics, origami, painting, pattern, plastic, plush sculpture drawing, tattoo, toy, video game, etc. The Imagenet-R dataset has undergone a deduction process that resulted in 30k images by removing 200 categories from the original ImageNet. Like the way of getting Split CIFAR-100, splitting ImageNet-R \cite{dualprompt} is also challenging due to the variance intra-class styles. The Split ImageNet-R benchmark randomly partitions the 200 classes into 10 tasks, each consisting of 20 classes. The dataset is then split into a training set and a test set under the portion of 8:2.\par
\textbf{CORe50}
The CORe50 dataset \cite{core50}, cited as a standard for continual object recognition, comprises 50 categories from 11 distinct domains. In the continual learning setting, incremental training is performed on data from eight domains (120K images), while the remaining (unseen) domains are used as test sets. Following the methodology of \cite{Sprompts} and \cite{18self} citations, we present average forward classification accuracy for CORe50 in this paper.
\begin{table}[!t]
	\centering
	\caption{Performance $A_a$ of CODA-P tested on the previous tasks after completing the training phase of each task on Split Cifar-100 dataset. `Tt' means ${t}$-th task.}
    \begin{tabular}{rrrrrrrrrr}
    \toprule
    \multicolumn{1}{l}{T1} & \multicolumn{1}{l}{T2} & \multicolumn{1}{l}{T3} & \multicolumn{1}{l}{T4} & \multicolumn{1}{l}{T5} & \multicolumn{1}{l}{T6} & \multicolumn{1}{l}{T7} & \multicolumn{1}{l}{T8} & \multicolumn{1}{l}{T9} & \multicolumn{1}{l}{T10} \\
    \midrule
    99.3  & 93.3  & 89.8  & 88.5  & 86.8  & 83.9  & 84.2  & 84.2  & 82.8  & 81.9  \\
          & 96.6  & 94.4  & 85.2  & 82.5  & 80.8  & 79.7  & 78.2  & 78.0  & 78.0  \\
          &       & 96.1  & 95.8  & 94.0  & 92.7  & 91.7  & 91.0  & 90.4  & 90.5  \\
          &       &       & 95.4  & 93.7  & 91.4  & 91.1  & 90.1  & 90.0  & 87.1  \\
          &       &       &       & 93.3  & 93.1  & 92.7  & 92.9  & 92.5  & 90.8  \\
          &       &       &       &       & 84.6  & 83.3  & 82.1  & 81.9  & 80.1  \\
          &       &       &       &       &       & 88.6  & 87.7  & 86.2  & 85.0  \\
          &       &       &       &       &       &       & 90.8  & 90.4  & 89.1  \\
          &       &       &       &       &       &       &       & 94.3  & 91.9  \\
          &       &       &       &       &       &       &       &       & 89.8  \\
    \bottomrule
    \end{tabular}%

	\label{tab:coda-cifar}%
\end{table}%

\begin{table}[!t]
	\centering
	\caption{Performance $A_a$ of CODA-P tested on the previous tasks after completing the training phase of each task on Split ImageNet-R dataset. `Tt' means ${t}$-th task.}
    \begin{tabular}{rrrrrrrrrr}
    \toprule
    \multicolumn{1}{l}{T1} & \multicolumn{1}{l}{T2} & \multicolumn{1}{l}{T3} & \multicolumn{1}{l}{T4} & \multicolumn{1}{l}{T5} & \multicolumn{1}{l}{T6} & \multicolumn{1}{l}{T7} & \multicolumn{1}{l}{T8} & \multicolumn{1}{l}{T9} & \multicolumn{1}{l}{T10} \\
    \midrule
    94.5  & 90.6  & 87.3  & 86.4  & 82.3  & 79.9  & 77.7  & 75.7  & 75.1  & 74.2  \\
          & 81.8  & 79.9  & 78.5  & 74.6  & 73.0  & 72.5  & 69.6  & 68.0  & 67.3  \\
          &       & 86.6  & 85.5  & 86.1  & 84.4  & 82.5  & 81.9  & 80.7  & 79.7  \\
          &       &       & 77.4  & 74.6  & 73.5  & 72.2  & 69.7  & 69.2  & 67.8  \\
          &       &       &       & 82.9  & 82.3  & 78.7  & 77.7  & 77.7  & 76.5  \\
          &       &       &       &       & 80.0  & 78.6  & 76.5  & 75.4  & 74.8  \\
          &       &       &       &       &       & 81.1  & 80.4  & 79.2  & 77.7  \\
          &       &       &       &       &       &       & 80.8  & 79.7  & 78.0  \\
          &       &       &       &       &       &       &       & 73.9  & 71.9  \\
          &       &       &       &       &       &       &       &       & 74.7  \\
    \bottomrule
    \end{tabular}%
	\label{tab:coda-R}%
\end{table}%

\textbf{DomainNet}
DomainNet \cite{domainnet} is a dataset consisting of 345 categories and approximately 600,000 images that can be utilized for domain adaptation and Domain Incremental Learning (DIL). The images from DomainNet are divided into six domains, with the DIL setup on this platform being identical to that of CaSSLe \cite{18self}. In accordance with \cite{Sprompts} and \cite{18self}, we present task-agnostic test average results for forward classification accuracy.

\subsection{Evaluation matrices}

For settings that involve task boundaries and where each task is associated with a test set, we utilize two widely-used metrics: Average Accuracy (higher values indicate better performance) and Forgetting (lower values indicate better performance) \cite{gradient2017}. After the training phase is completed for task $t$, we assess the current learner's performance on all previous tasks using their respective test sets. Let ${{a}_{t,j}} \in \left[ {0,1} \right]$ denote the accuracy achieved on task $j$-th test set after training on task $t$. The Average Accuracy (${A_a}$) measures the average value of ${{a}_{t,j}}$ as:\par

\begin{equation}
	{{A_a} = \frac{1}{t}\sum\limits_{j = 1}^t {{{a}_{t,j}}} }
 \label{eq:aa}
\end{equation}

And ${{f}_{t,j}}$ denotes the forgetting transfer calculated on task $j$-th test set after training on task $t$ which is define as:\par

\begin{equation}
	{f_j^t = \begin{array}{*{20}{c}}
			{\max } \\ 
			{i \in \left\{ {1, \cdots t - 1} \right\}} 
		\end{array}{a_{i,j}} - {a_{t,j}} }
\end{equation}

Hence, the Forgetting (${F_t}$) measures the average value of ${f_j^t}$ as: \par

\begin{equation}
	{{F_t} = \frac{1}{{t - 1}}\sum\limits_{j = 1}^{t - 1} {f_j^t}}
 \label{eq:ft}
\end{equation}

\section{Implementation details and training schedule}

We apply task-agnostic prompts in the first 2 blocks like DualPrompt \cite{dualprompt} inspired by Complementary Learning Systems
(CLS) \cite{cls1995there,cls2016learning}  and the proposed modulated instance-specific prompts in the third to fifth blocks like DualPrompt. As for the prompting function to combine the modulated instance-specific prompts with input embedding, we employ the Prefix Tuning (Pre-T) referring to the DualPrompt \cite{dualprompt}.\par

While considering applying prompts in different blocks, the PC establishes an independent PRM for each block and assigns a shared codebook and a shared PMM to all the PRMs of all blocks. Accordingly, the instance-specific prompts are divided into two parts $\hat P_{x_i^t}^K$ and $\hat P_{x_i^t}^V$ for key and values of ${M_{SA}}$ layer in each VIT block, respectively. $\hat P_{x_i^t}^K$ and $\hat P_{x_i^t}^V$ are all with the size of ${\mathbb{R}^{{n} \times L}}$. Specifically, the application of a prompting function can be interpreted as modifying the inputs of the ${M_{SA}}$ layers \cite{attention2017}. Let $h \in {\mathbb{R}^{D \times L}}$ denote the input to the ${M_{SA}}$ layer, and let ${h^Q}$, ${h^K}$, and ${h^V}$ represent its input query, key, and values, respectively. Hence, the corresponding prompts are combined by Pre-T in each corresponding ${M_{SA}}$ layer as follows:

\begin{equation}
	{{f_{p}}({p^\ell},{h^\ell}) = {M_{SA}}({h^{Q,\ell}},[\hat P_{x_i^t}^{K,\ell},{h^{K,\ell}}],[\hat P_{x_i^t}^{V,\ell},{h^{V,\ell}}])}
\end{equation}
where $\ell$ means $\ell$-th block. 

The model is trained for 5 epochs for each task by iteratively minimizing the loss ${\mathcal L}$. Adam optimization is employed for training with a batch size of 64 on NVIDIA A100 GPU and a momentum of 0.9. The initial learning rate is set as $0.007$.

\section{Hyperparameter analyses}

\begin{figure}[!t]
	\centerline{\includegraphics[width=1\columnwidth]{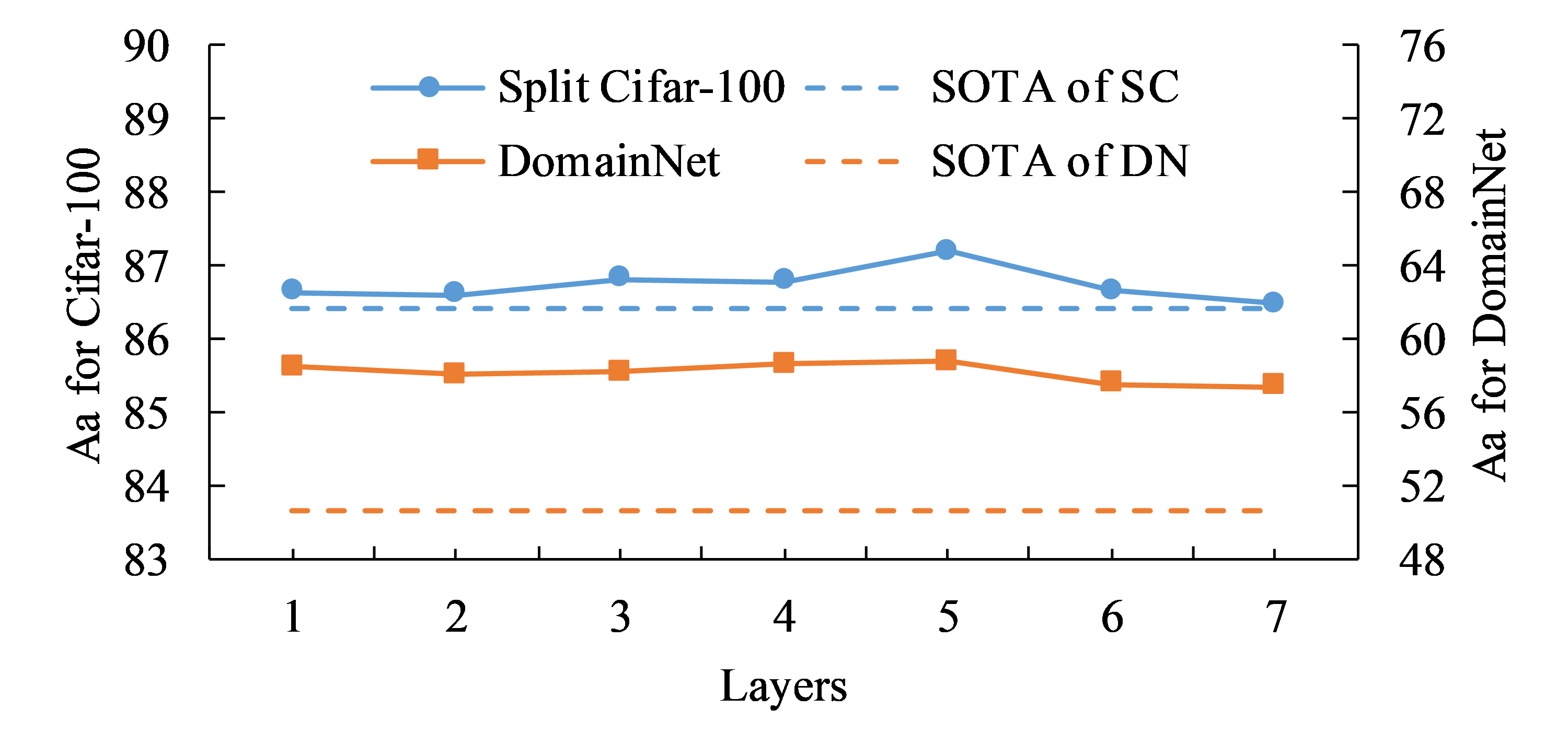}}
	\caption{The performance with regard to the number of attention layers of PGM. }
	\label{fig:layers_C}
\end{figure}

\begin{figure}[!t]
	\centering{ \subfigure[]{\includegraphics[width=1\columnwidth]{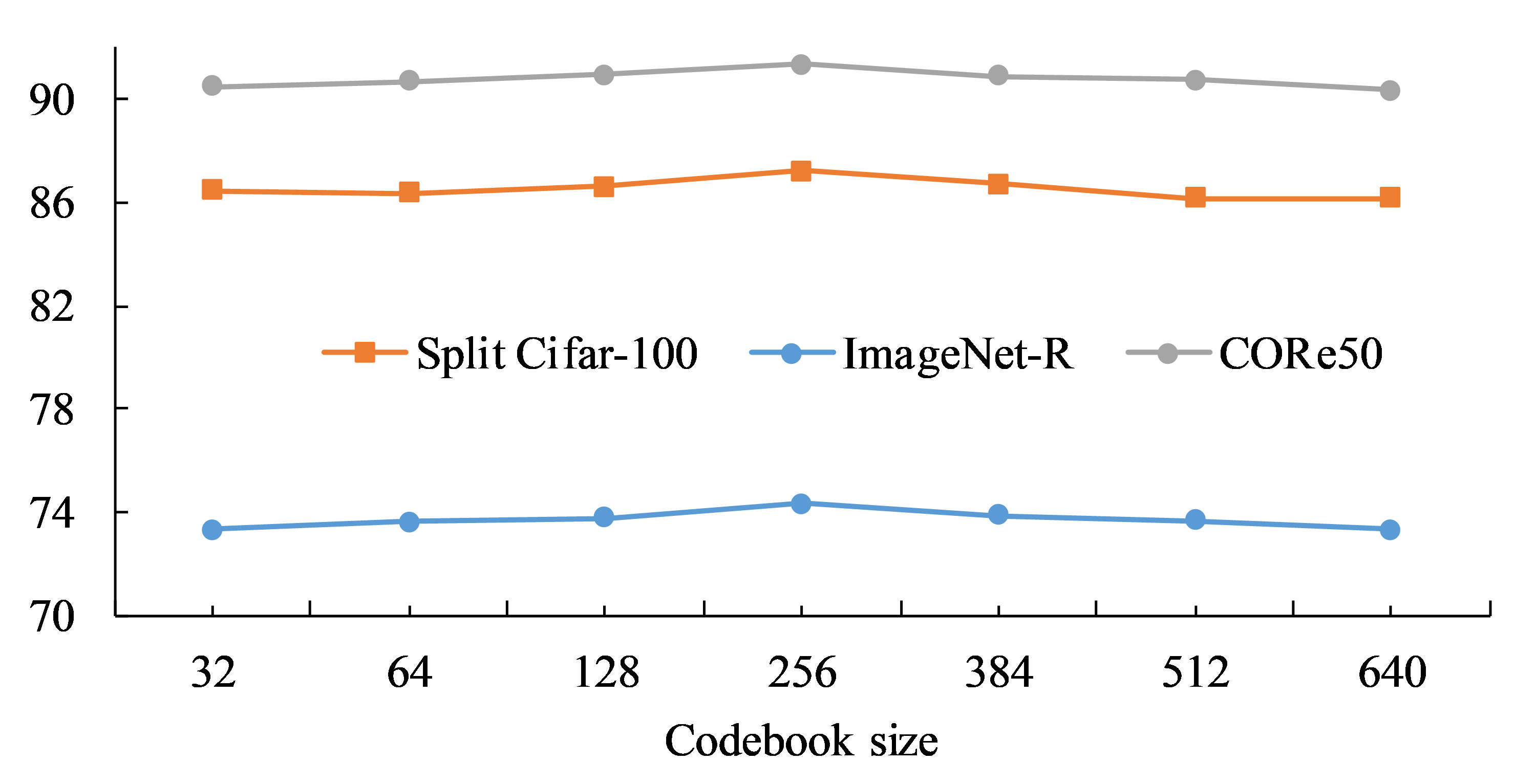}}
		\subfigure[]{\includegraphics[width=.9\columnwidth]{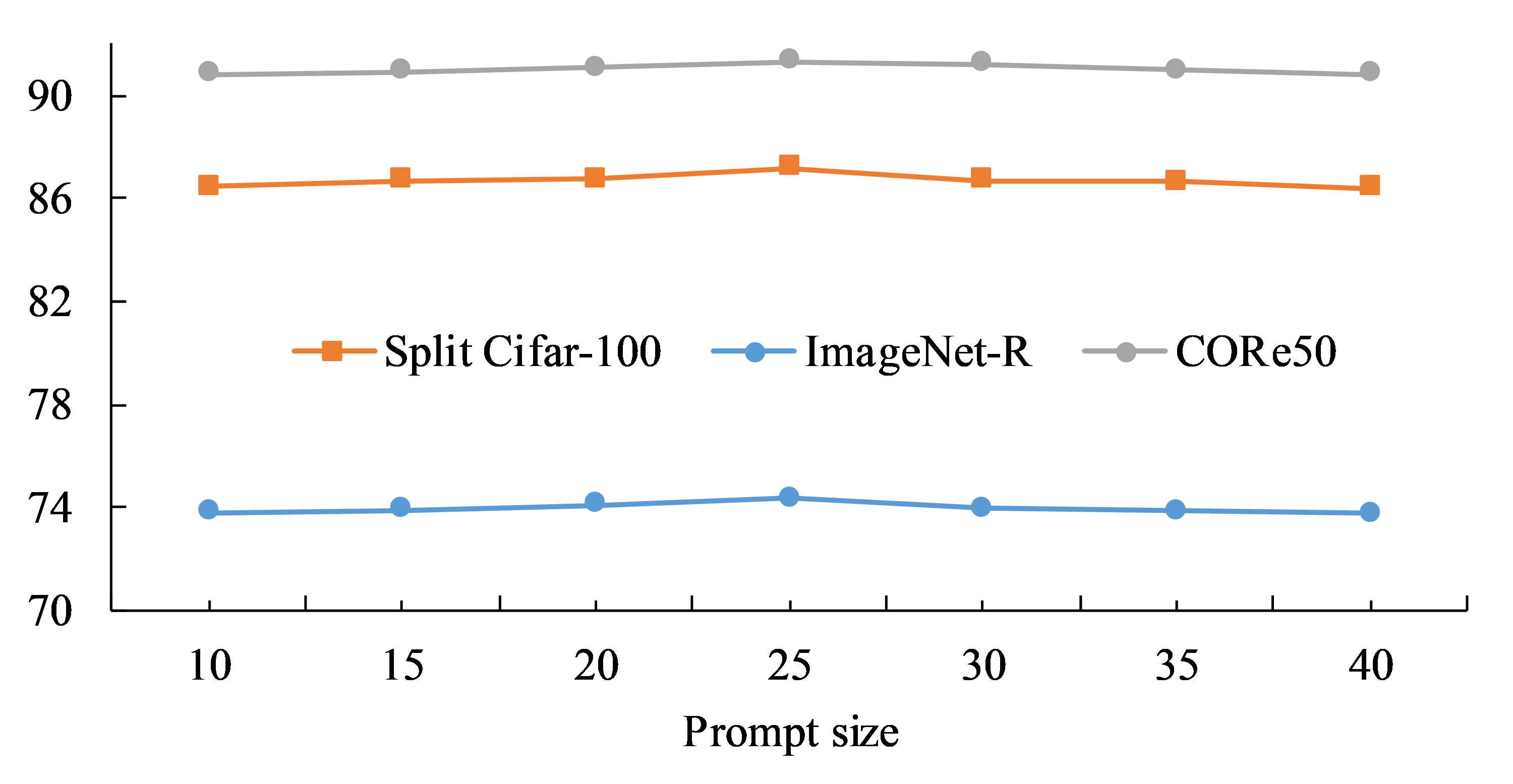}}
  \subfigure[]
  {\includegraphics[width=.9\columnwidth]{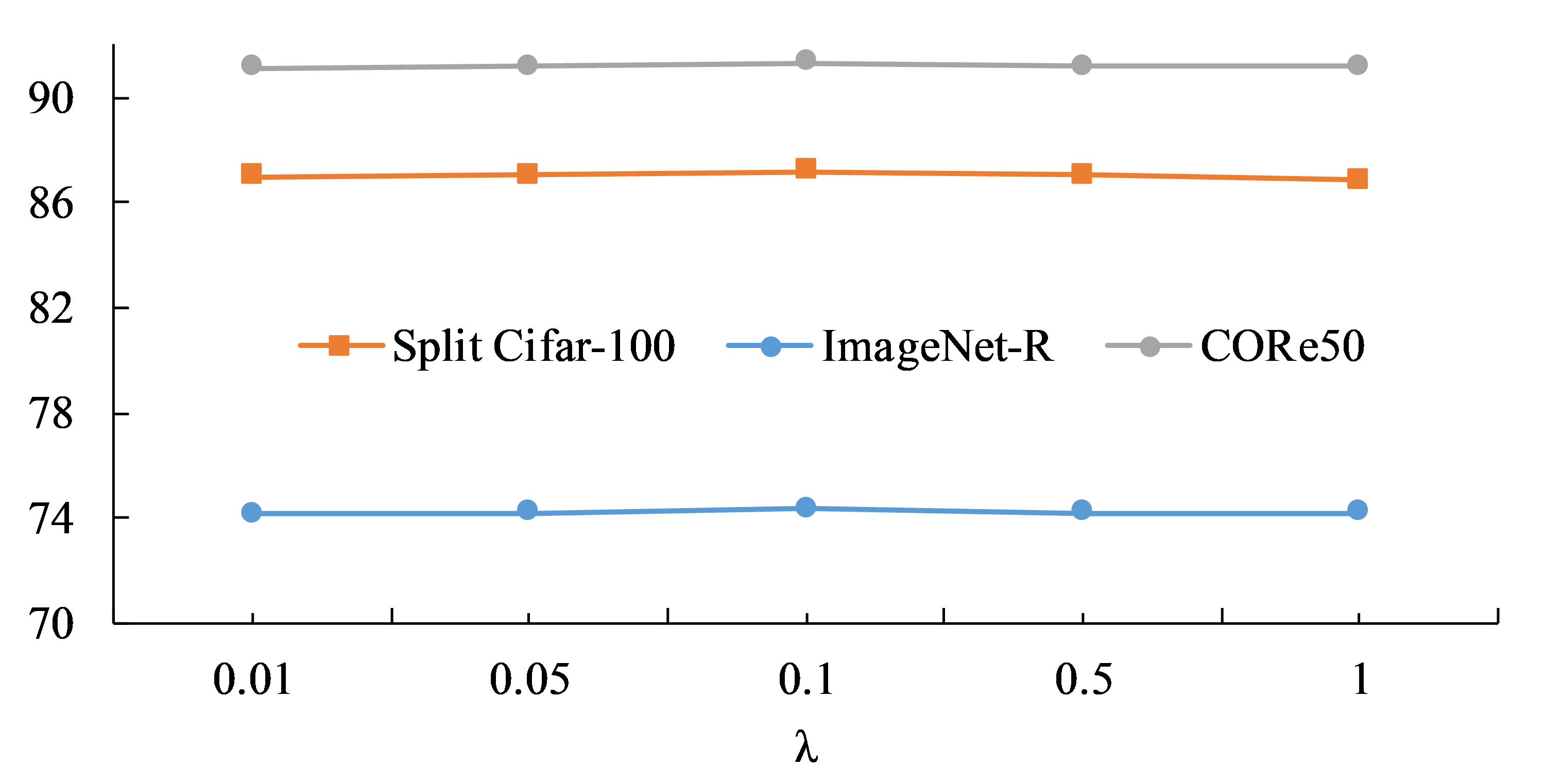}}
  }
\caption{The performance with regard to the proposed codebook size, prompt size, and the weight $\lambda$.}
\label{fig:other_analyse}
\end{figure}




The attention computation employs several ViT blocks to explore the relationships between each other. Fig. \ref{fig:layers_C} reports the performance with regard to the number of attention layers of PGM, the performance shows small fluctuations when the layer number is less than 5 and drops with the number further increasing.\par

Besides, we also give the performance results with regard to the proposed codebook size, prompt size, and the weight $\lambda$ in Fig. \ref{fig:other_analyse}. The results also demonstrate the stability of the proposed PC with small fluctuations concerning different parameters.

\section{Detailed experiment results of CODA-P}

We train CIFAR-100 for 20 epochs, and ImageNet-R for 50 epochs according to the original paper. Hence, it requires almost 4 times the training time of DualPrompt (5 epochs) on CIFAR-100. Tab .\ref{tab:coda-cifar} and Tab. \ref{tab:coda-R} give the performance of CODA-P tested on the previous tasks after completing the training of each task on Split CIFAR-100 dataset and Split ImageNet-R dataset, respectively. The forgetting results of CODA-P in the original paper are different from the experimental results, we report the corresponding performances of $A_a$ and $F$ for CODA-P according to the above tables referring to the eq. \ref{eq:aa} and \ref{eq:ft}.





\bibliographystyle{ACM-Reference-Format}
\bibliography{refPC}

\end{document}